\documentclass[conference]{IEEEtran}
\IEEEoverridecommandlockouts
\usepackage{cite}
\usepackage{amsmath,amssymb,amsfonts}
\usepackage{algorithmic}
\usepackage{graphicx}
\usepackage{textcomp}
\usepackage{xcolor}
\def\BibTeX{{\rm B\kern-.05em{\sc i\kern-.025em b}\kern-.08em
    T\kern-.1667em\lower.7ex\hbox{E}\kern-.125emX}}
\begin{document}

\title{MissionGPT: Mission Planner for Mobile Robot based on Robotics Transformer Model\\
}

\author{\IEEEauthorblockN{Vladimir Berman}
\IEEEauthorblockA{\textit{Skolkovo Institute of} \\
\textit{Science and Technology} \\
Moscow, Russia \\
Vladimir.Berman@skoltech.ru}
\and
\IEEEauthorblockN{Artem Bazhenov}
\IEEEauthorblockA{\textit{Skolkovo Institute of} \\
\textit{Science and Technology} \\
Moscow, Russia \\
Artem.Bazhenov@skoltech.ru}
\and
\IEEEauthorblockN{Dzmitry Tsetserukou}
\IEEEauthorblockA{\textit{Skolkovo Institute of} \\
\textit{Science and Technology} \\
Moscow, Russia \\
d.tsetserukou@skoltech.ru}
}

\maketitle

\begin{abstract}
This paper presents a novel approach to building mission planners based on neural networks with Transformer architecture and Large Language Models (LLMs). This approach demonstrates the possibility of setting a task for a mobile robot and its successful execution without the use of perception algorithms, based only on the data coming from the camera. In this work, a success rate of more than 50\% was obtained for one of the basic actions for mobile robots. The proposed approach is of practical importance in the field of warehouse logistics robots, as in the future it may allow to eliminate the use of markings, LiDARs, beacons and other tools for robot orientation in space. In conclusion, this approach can be scaled for any type of robot and for any number of robots.
\end{abstract}


\section{Introduction}
Over the past decade, the mobile robots’ sector has experienced significant global growth. Industrial mobile robots are becoming increasingly advanced to enhance autonomy and efficiency across various industries \cite{b1}. These robots are outfitted with sophisticated sensors like Light Detection and Ranging (LiDAR), stereo cameras, Inertial Measurement Units (IMU), and global or indoor positioning systems to gather environmental data and make informed decisions. Complex algorithms enable these robots to plan paths, avoid obstacles, and execute tasks. Additionally, fleets of autonomous mobile robots are often integrated with cloud-based technologies, facilitating remote monitoring and control for greater flexibility and scalability.
Path planning is essential for mobile robot navigation, requiring the robot to move from one point to another while avoiding obstacles and meeting constraints such as time, energy autonomy, and safety for human operators and transported cargo \cite{b2,b3}. Navigation in mobile robotics remains a highly researched area, focusing on two main categories: classical and heuristic navigation \cite{b4,b5}. Classical approaches, known for their limited intelligence, include algorithms like cell decomposition, the roadmap approach, and artificial potential fields (APF). More intelligent heuristic approaches incorporate computational intelligence components such as fuzzy logic, neural networks, and genetic algorithms \cite{b6}.
\begin{figure}[htp]
    \centering
    \includegraphics[width=9cm]{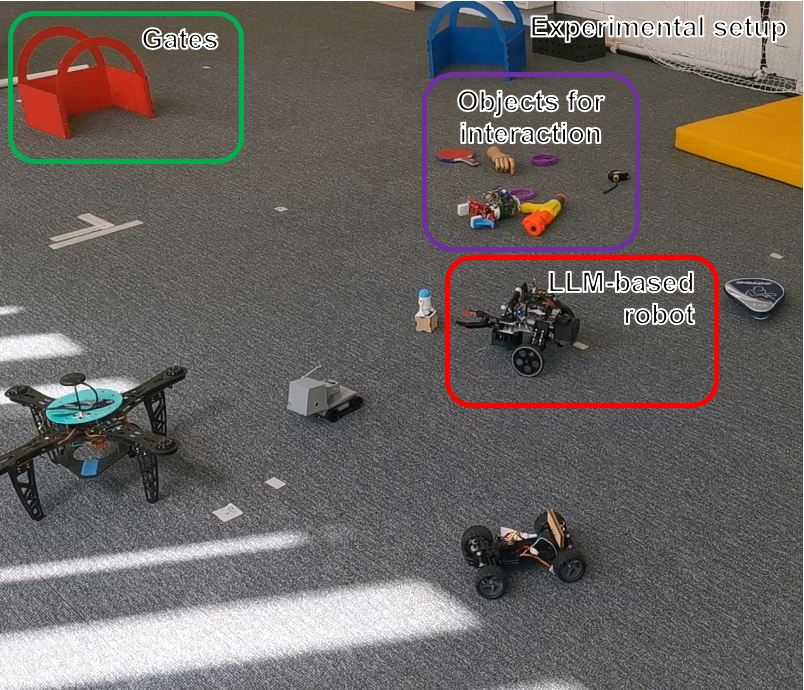}
    \caption{Experimental setup view}
    \label{fig:setup}
\end{figure}
Researchers also explore solutions using algorithms like particle swarm optimization, the Firefly algorithm, and the artificial bee colony algorithm. Hybrid algorithms, which combine classical and heuristic methods, offer superior performance, particularly in complex and dynamic environments \cite{b7}.
On the other side developing an end-to-end robotics transformer-based model for mobile robot control represents an innovative frontier, merging artificial intelligence and robotics \cite{b8,b9}. Such models enhance efficiency by streamlining control processes, adapting to diverse environments and tasks, and generalizing learning across scenarios. Their deployment holds immense potential for real-world applications across industries, from warehouse automation to healthcare assistance. These models enable autonomous exploration, empowering robots to navigate complex terrains and accomplish tasks with minimal human intervention. With scalability and robustness, they promise to revolutionize industries and improve human lives by enabling safer, more efficient, and autonomous robotic systems \cite{b10}.

\section{Related Works}

\subsection{RT-1}

The RT-1 model introduced in this study represents a cutting-edge advancement in the realm of real-time robotic control systems \cite{b11}. With a focus on achieving fast and consistent inference speed, the model integrates a sophisticated Transformer architecture with key components such as EfficientNet \cite{b12}, FiLM conditioning, TokenLearner, and Transformer. This unique combination enables the model to efficiently process both image and natural language inputs, facilitating the generation of precise and timely actions for robotic manipulation tasks \cite{b13,b14}.
One of the key strengths of the RT-1 model lies in its capacity to absorb data from various sources, including real-world and simulation data. This capability enhances the model's generalization across different tasks, objects, and environments, enabling it to adapt and perform effectively in diverse settings. Moreover, the model's architecture allows for closed-loop control, enabling it to command actions at a rapid pace of 3 Hz until a termination signal is received or a predefined time step limit is reached. This real-time control mechanism ensures that the model can respond swiftly to changing conditions and execute tasks with precision and efficiency.
In addition to its technical capabilities, the RT-1 model offers a scalable and adaptable solution for a wide range of robotic manipulation tasks. Its efficient processing of high-dimensional inputs, coupled with its ability to generate actions in real-time, makes it well-suited for applications that require fast and accurate decision-making.
Overall, the RT-1 model represents a significant contribution to the field of robotics and artificial intelligence, offering a sophisticated yet practical solution for real-time control in complex manipulation tasks. Its innovative architecture, efficient processing capabilities, and robust performance make it a valuable tool for researchers and practitioners seeking to enhance the capabilities of robotic systems in various domains.

\subsection{RT-2}

The work delves into the intricate details of model architectures and training strategies in the realm of robotic control \cite{b15}. It emphasizes the significance of co-fine-tuning over fine-tuning for achieving better generalization in model performance. The RT-2-PaLI-X and RT-2-PaLM-E models are specifically highlighted for their successful co-fine-tuning with robotic data, showcasing improved generalization capabilities \cite{b16}.
Furthermore, the evaluation scenarios outlined in the research focus on assessing the reasoning, symbol understanding, and human recognition abilities of the RT-2 model. By providing a comprehensive set of evaluation instructions and training parameters, the study offers valuable insights for researchers and practitioners in the field of vision-language-action models for robotic control \cite{b17}.
It underscores the importance of thoughtful model design and training methodologies in enhancing the performance and generalization capabilities of robotic systems, paving the way for more efficient and effective human-robot interactions in real-world environments.

\subsection{PRIMAL}

This work presents a transformative approach to multi-robot path planning by integrating transformer structures into policy neural networks \cite{b18}. This innovative framework combines imitation learning and reinforcement learning techniques to enhance the robots' ability to navigate efficiently and collision-free in dense environments without the need for inter-robot communication. The study emphasizes decentralized solutions, focusing on coordination and decentralized path planning for large-scale systems \cite{b19}. The introduction of the transformer structure into policy neural networks marks a significant advancement, enabling the networks to extract features effectively and guide robots in complex environments. Through a detailed exploration of the observation space, action space, and reward structure, the study provides a comprehensive understanding of the learning environment for multi-robot path planning.
Furthermore, the proposed transformer-based imitation reinforcement learning method is detailed, encompassing the transformer-based policy network and the imitation reinforcement learning framework. The deep neural network based on the transformer structure is designed to map current observations and actions, facilitating policy approximation in a partially observable grid world. The framework allows agents to learn from expert demonstrations, enhancing their ability to collaborate efficiently on tasks \cite{b20}.
The results demonstrate significant performance improvements, indicating the potential of the transformer-based approach in revolutionizing multi-robot path planning and coordination in complex environments.

\subsection{PERACT}

The PERACT framework is a cutting-edge approach designed for language-conditioned behavior cloning, specifically tailored for 6-DoF manipulation tasks \cite{b21}. It distinguishes itself by leveraging voxelized observations and actions, employing a single multi-task Transformer model trained on a comprehensive set of 18 RLBench tasks and 7 real-world tasks. Noteworthy is the framework's superior performance compared to traditional image-to-action agents and 3D ConvNet baselines, showcasing its effectiveness across a range of tabletop tasks. For example, the meticulously reported success rates of multi-task agents trained on the 18 tasks underscore the framework's proficiency in task execution. The demonstrations within the study involve expert actions paired with English language goals, illustrating the framework's adeptness in interpreting and executing tasks based on natural language instructions.
The considered works have in their basis the same idea - the use of transformer architecture. Some of the works use this model in its pure form, the other part uses neural networks pre-trained on a large amount of data. The latter approach has more prospects, as it uses the knowledge embedded in natural language, so for the study it was decided to use a similar approach in application to mobile robots.

\section{System Overview}
To verify the performance of the proposed solution, a test bench KabutoBot with hardware and software parts has been developed.

\subsection{Hardware system}
\begin{figure}[htp]
    \centering
    \includegraphics[width=9cm]{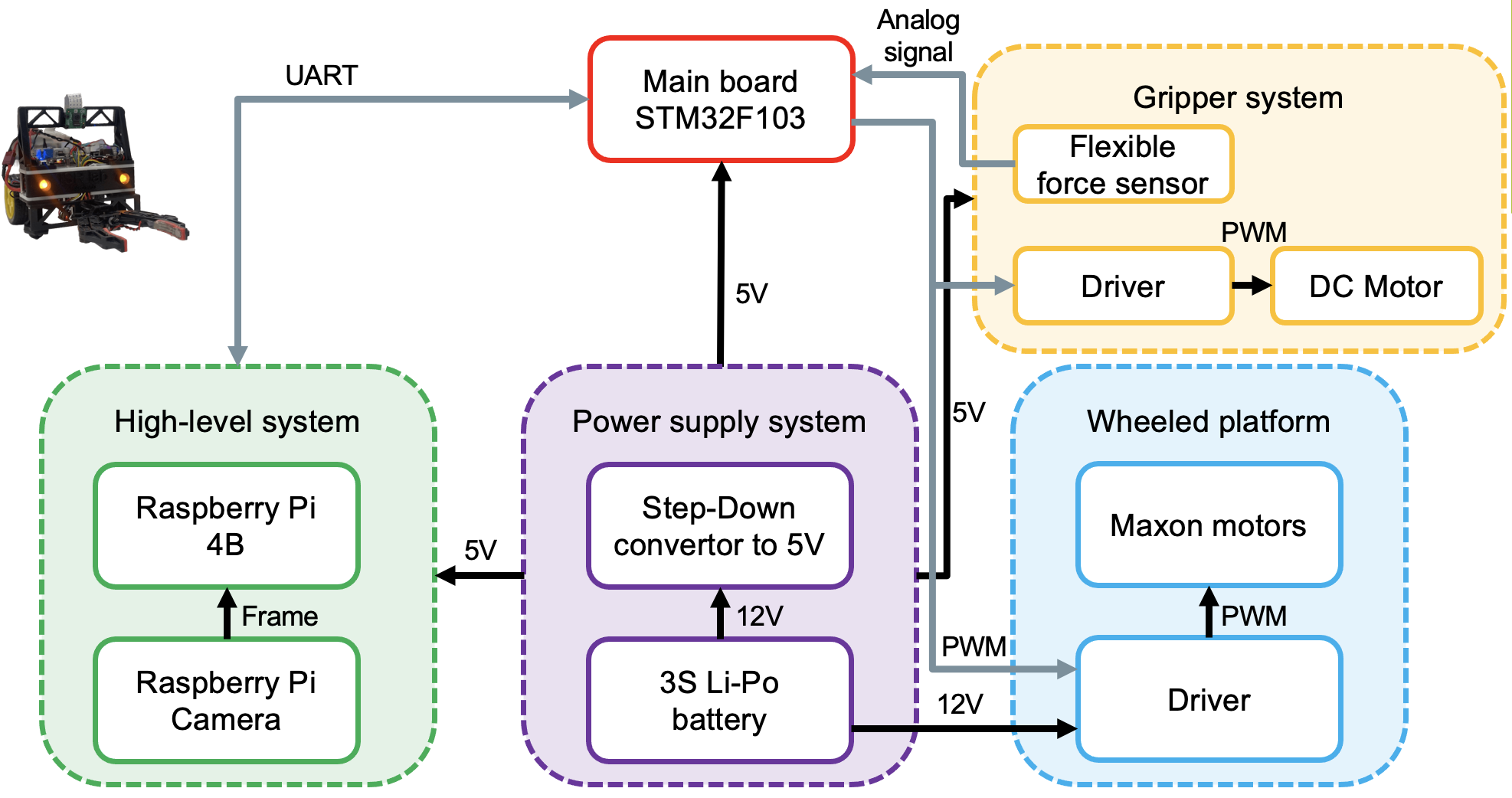}
    \caption{Hardware system architecture}
    \label{fig:hard}
\end{figure}

The platform comprises five principal components: a high-level system incorporating a Raspberry Pi 4B on-board computer and an RGB camera, a main board based on an STM32 microcontroller, a power system including a battery and voltage converter, a gripper system equipped with a gripper and DC motor utilizing flexible sensors for compression force detection, and a wheel platform driven by Maxon motors (Fig. \ref{fig:hard}).
\subsection{Software system}
\begin{figure}[htp]
    \centering
    \includegraphics[width=9cm]{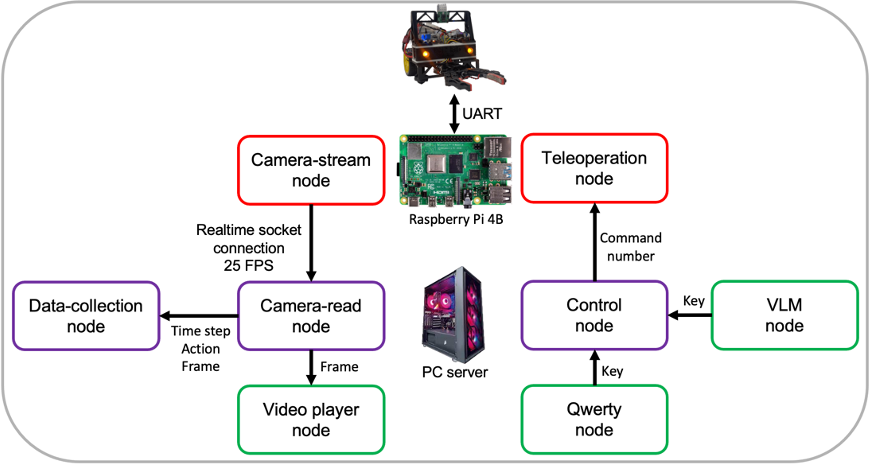}
    \caption{Software system architecture}
    \label{fig:sa}
\end{figure}

The connection between computers bases on usage of the ROS2 Hummble framework. For system control was developed several ROS2 nodes for different purposes: data transferring, dataset collection and markup, command sending and converting (Fig. \ref{fig:sa}).
 The 160 x 120 resolution image is transmitted from the robot to the server at a frequency of 25 Hz, and a socket connection using UDP protocol and message queues is established for transmission between computers, which reduces the latency to 30 milliseconds. The Camera-read node sends data to the Video player node for display on the user's screen and to the Data-collection node for writing images to the server's memory. The Control node sends commands to the Teleoperation node based on the received robot state; the data in the Control node can come from the keyboard during data collection or directly from the model during operation.

\subsection{Dataset specification}

\begin{figure}[htp]
    \centering
    \includegraphics[width=9cm]{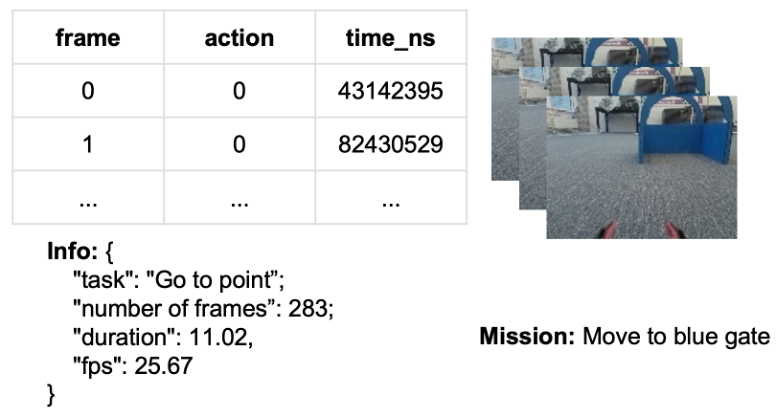}
    \caption{Data sample example}
    \label{fig:dataset}
\end{figure}

Dataset consists of 750 samples: 150 ``Go to point'' task and 600 ``Pick and place'' task. Each sample includes 100-1500 frames, task description, information file and mapping file between each frame and action performed in this time moment. Dataset was collected with 25 fps; total data volume is 5 hours (4,6 GB). Example of data sample is presented in (Fig. \ref{fig:dataset}).
The action space consists of a set of numbers from 0 to 9, where each number corresponds to the robot's action at a single moment in time (100 ms). Actions include the robot's movement in different directions and opening/closing of the gripper.
\begin{table}[htbp]
\caption{KabutoBot Action Space}
\begin{center}
\begin{tabular}{|c|c|}
\hline
{\textbf{Action}}&{\textbf{Number}} \\
\hline
{Move forward}&{0} \\
\hline
{Move back}&{1} \\
\hline
{Move right}&{2} \\
\hline
{Move left}&{3} \\
\hline
{Move forward+right}&{4} \\
\hline
{Move forward+left}&{5} \\
\hline
{Move back+right}&{6} \\
\hline
{Move back+left}&{7} \\
\hline
{Open gripper}&{8} \\
\hline
{Close gripper}&{9} \\
\hline
\end{tabular}
\label{tab1}
\end{center}
\end{table}

\subsection{Model architecture}

During research two different approaches was implemented: encoder only neural network architecture and LLM with visual input. Important point is that models were pretrained on web scale volume datasets, so this information and inner dependencies provide generalization to task performing.

\subsubsection{Encoder-only model}
\begin{figure}[htp]
    \centering
    \includegraphics[width=9cm]{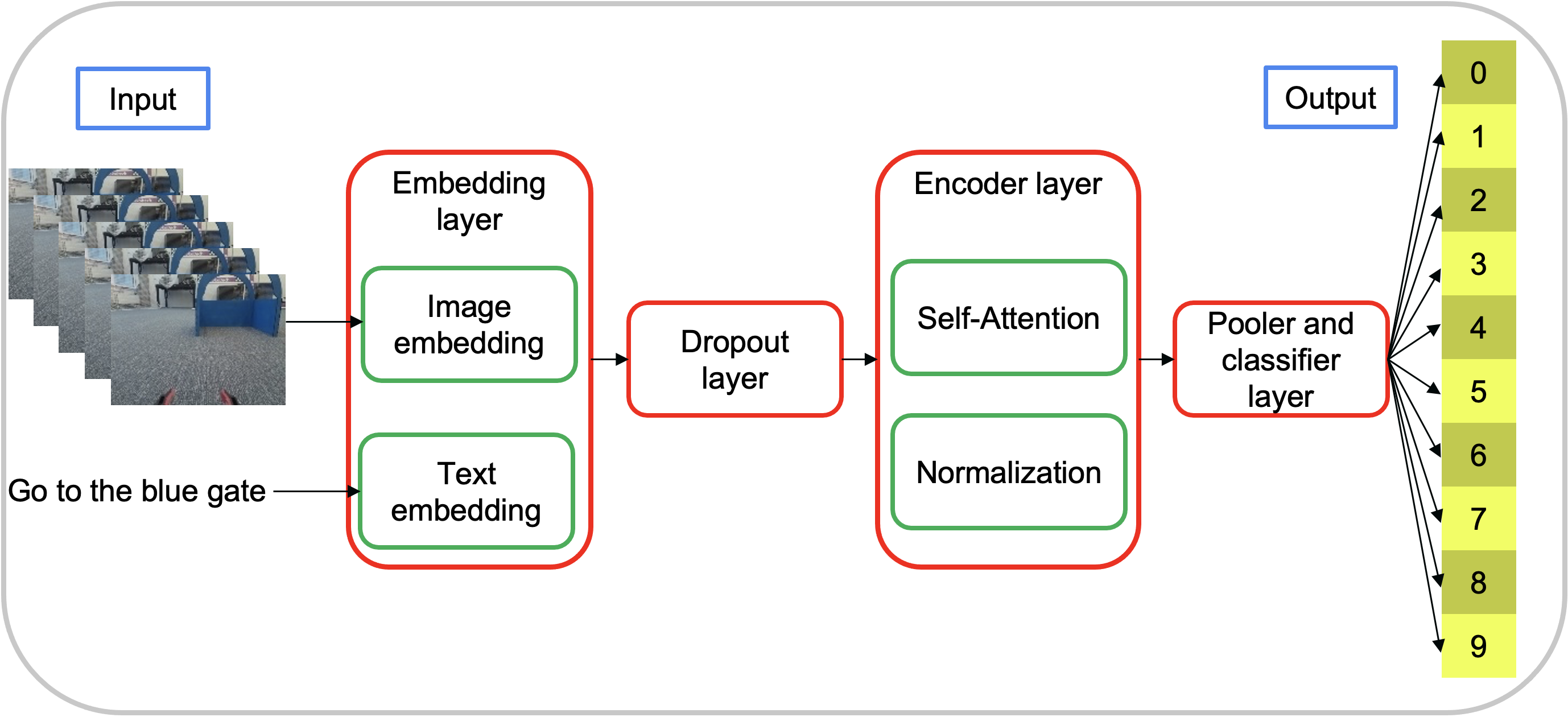}
    \caption{Encoder-only model scheme}
    \label{fig:encoder}
\end{figure}

Encoder-Only architecture presented in Fig. \ref{fig:encoder}. The proposed model architecture offers a robust framework tailored for classification tasks, seamlessly integrating both textual and visual information. Total number of parameters is 87.4M. Embracing an encoder-only design, the model efficiently processes input sequences comprising text and photo embeddings in a unified manner. Beginning with initialization, the model parameterizes tokenization functions, embedding matrices, and classification layers, setting the stage for comprehensive processing. Textual and photo inputs are independently tokenized and then converted into embeddings, subsequently merged to form a cohesive input representation. This integration enables the model to capture nuanced relationships between textual and visual elements, laying the groundwork for enhanced feature extraction and classification performance.
Following the initial processing stage, the model progresses to encoder-based feature extraction, where informative features are extracted from the combined embeddings. Leveraging global average pooling, the model efficiently aggregates these features, facilitating effective dimensionality reduction while preserving critical information. Subsequently, a fully connected layer applies classification, generating class probabilities through softmax activation. This architecture not only accommodates multimodal inputs but also fosters seamless fusion and processing, culminating in accurate and interpretable classification outcomes. By capitalizing on the synergies between text and visual data, the proposed model architecture offers a promising avenue for robust and interpretable classification across diverse application domains. The rationale for using the encoder architecture is to be able to interpret the task of predicting the robot's action at the next point in time as a classification. The input is 4 previous frames and the current camera image, and the output of the fully-connected layer using the ArgMax function is one of the 10 actions shown in Table \ref{tab1}.

\subsubsection{Encoder-decoder model}
\begin{figure}[htp]
    \centering
    \includegraphics[width=9cm]{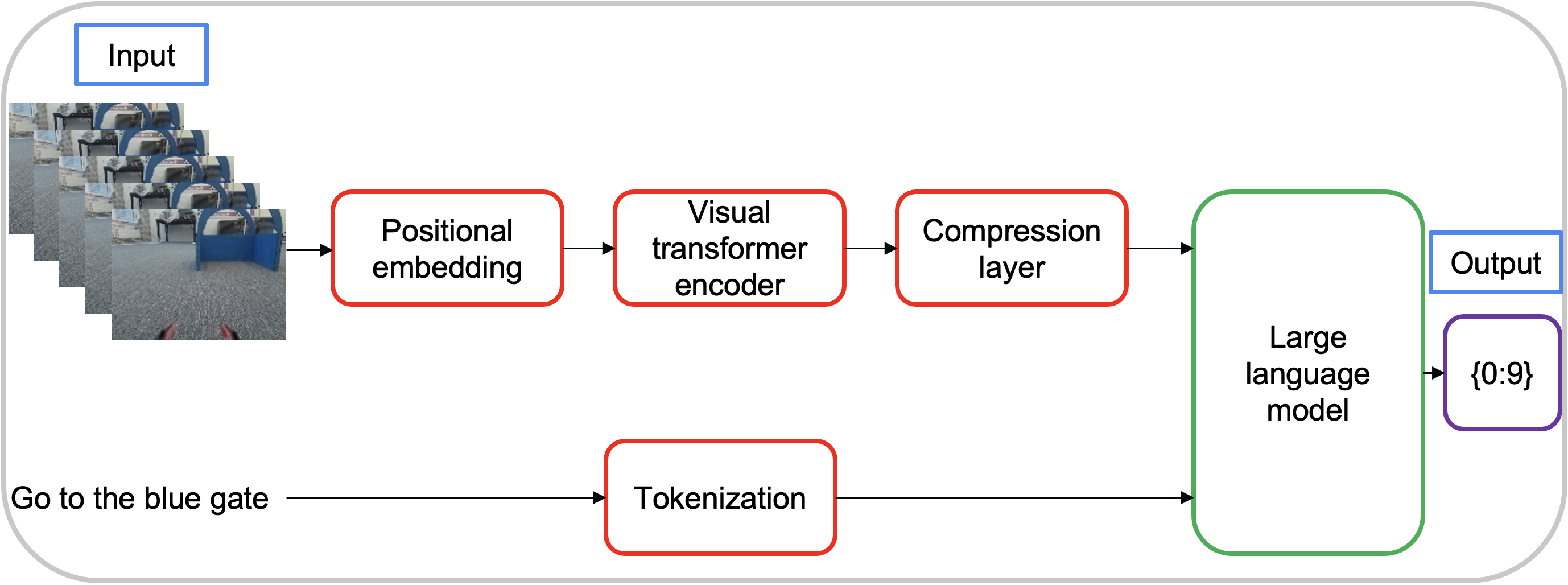}
    \caption{ViLT model scheme}
    \label{fig:vilt}
\end{figure}

Other approach is utilizing encoder-decoder transformer architecture. For this case LLMs with visual input (ViLT) are suitable due to their large capacity. System architecture is presented in Fig. \ref{fig:vilt}.
The neural network model integrates techniques in computer vision and natural language processing for the fusion of photo and text inputs. At its core lies a transformer-based architecture, specifically designed for multimodal tasks. Initially, the input photos are concatenated and scaled, followed by partitioning to capture diverse visual features effectively. Positional embedding ensures the model retains spatial information across partitions and the entire frame. Leveraging CLIP-ViT-L/14, the vision transformer encodes the visual information, capturing intricate relationships between images and extracting high-level features crucial for understanding context.
After encoding the visual data, a shared compression layer, employing average pooling, condenses the representation while preserving essential features. Simultaneously, the textual input undergoes embedding and concatenation with the visual representation, facilitating seamless fusion of multimodal information. The combined sequence is then passed to Vicuna-13B, a language and vision model adept at generating coherent predictions from multimodal inputs. Vicuna-13B contextualizes the fused information, leveraging its understanding of both images and text to generate a token that encapsulates the essence of the input. In parallel, an encoder-decoder architecture operates on source and target sequences, further enriching the model's capabilities. The encoder tokenizes and encodes the source sequence, while the decoder leverages positional encoding and transformer layers to generate predictions for the target sequence. Finally, a linear projection layer followed by softmax activation produces the predicted sequence. This model seamlessly integrates the power of transformers, vision transformers, and language and vision models, enabling sophisticated fusion of photo and text inputs to generate meaningful predictions. The idea behind this approach is to generate the next step based on a series of data and a verbal description of the problem. The input of the model is also multimodal data, the series of images is split into frames, each of which passes through a separate layer, further connecting to the embedding of the whole concatenated image. The output is 1 new token, which, as in the previous method, is interpreted as a command number (Table \ref{tab1}) for the robot at the next instant of time.

\subsubsection{Experimental setup}

The experimental setup involves the use of KabutoBot within an environment containing three endpoints identified by the colors red, green, and blue. The setup includes various objects such as toys, instruments, and figures. The primary tasks in this experiment are ``Go to point'' and ``Pick and place''. In the ``Go to point'' task, the robot is programmed to move from its current position to a specified colored endpoint, assessing its navigational accuracy. In the ``Pick and place'' task, robot searches for a designated object, grasps it, and transports it to a predetermined endpoint, evaluating its object recognition and manipulation capabilities. This setup allows for the systematic testing of mission planners' navigation and object interaction performance under controlled conditions (Fig. \ref{fig:setup}).

\section{Experimental Results}

\subsection{Experiment 1, ``Go to point'' task}
\begin{figure}[htp]
    \centering
    \includegraphics[width=9cm]{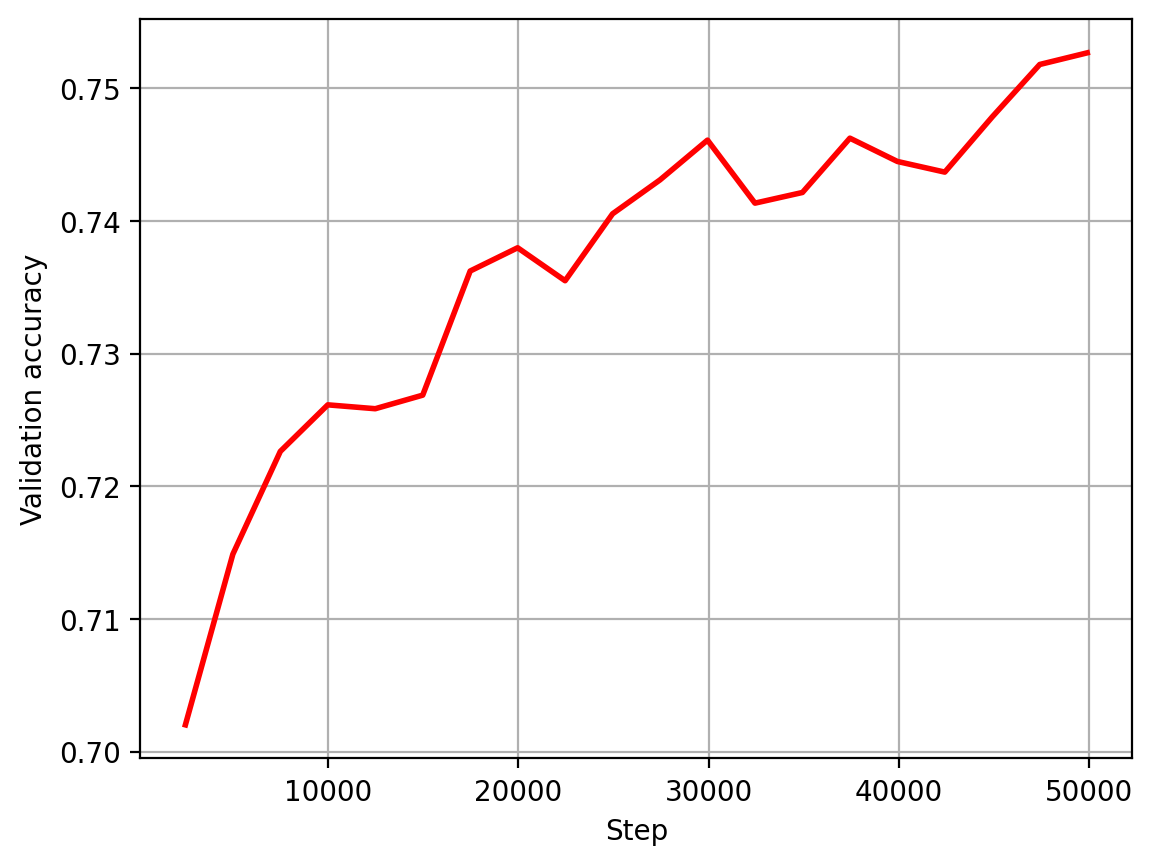}
    \caption{Validation plot}
    \label{fig:go}
\end{figure}

\begin{figure}[htp]
    \centering
    \includegraphics[width=9cm]{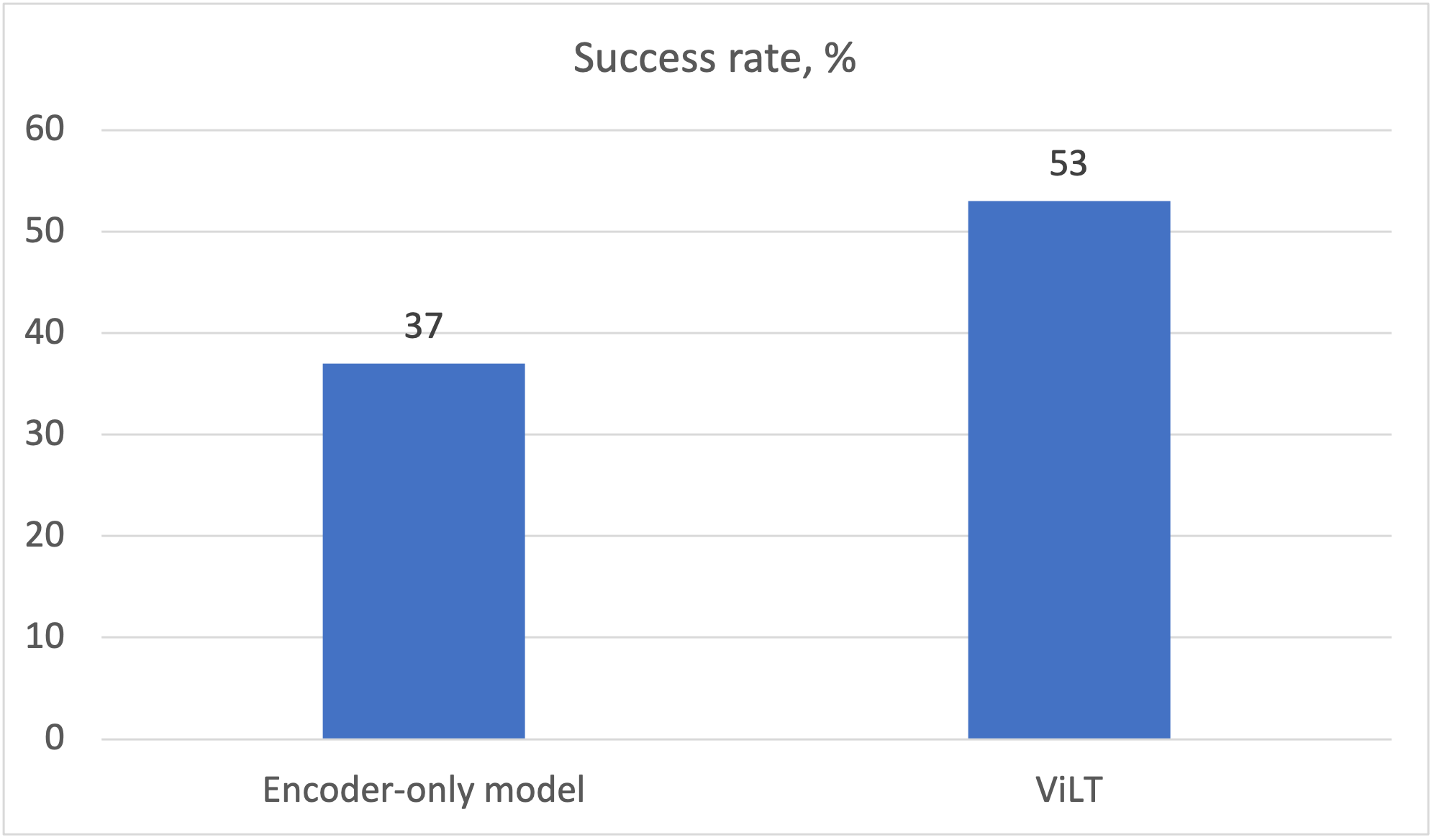}
    \caption{Success rate}
    \label{fig:plot_go}
\end{figure}

For this experiment, 150 recordings of video from the robot's camera and the actions it performs were collected. The video recordings were spun at 25 frames per second and 160 x 120 px resolution. Each sample consists of 600-800 frames and an action number for each frame. Successful completion of the task is considered to be receiving a command from the operator, determining your current position relative to the object to be driven to, moving and stopping near or inside the object. Results of validation during learning are presented in Fig. \ref{fig:go}.
The accuracy graph shows that the model was able to find certain dependencies in the training sample. After training, 30 test runs were conducted, 10 to gates of each color: blue, green and red. The final success results for the two models are shown in the graph (Fig. \ref{fig:plot_go}) and Table \ref{tab2}.
As shown in Table \ref{tab2}, there is no correlation between gate color and mission success, which means that the correct representation of the object and its presence in the dataset is important for the model.
\begin{table}[htbp]
\caption{Results of ``Go to point'' task for different gates}
\begin{center}
\begin{tabular}{|c|c|c|}
\hline
{\textbf{Gate}}&\multicolumn{2}{|c|}{\textbf{Race success rate}} \\
\cline{2-3}
{\textbf{color}}&{\textbf{Encoder-only model}}&{\textbf{ViLT}} \\
\hline
{Red}&{4/10}&{5/10}\\
\hline
{Blue}&{4/10}&{6/10} \\
\hline
{Green}&{3/10}&{5/10} \\
\hline
\end{tabular}
\label{tab2}
\end{center}
\end{table}

\subsection{Experiment 2, ``Pick and place'' task}
\begin{figure}[htp]
    \centering
    \includegraphics[width=9cm]{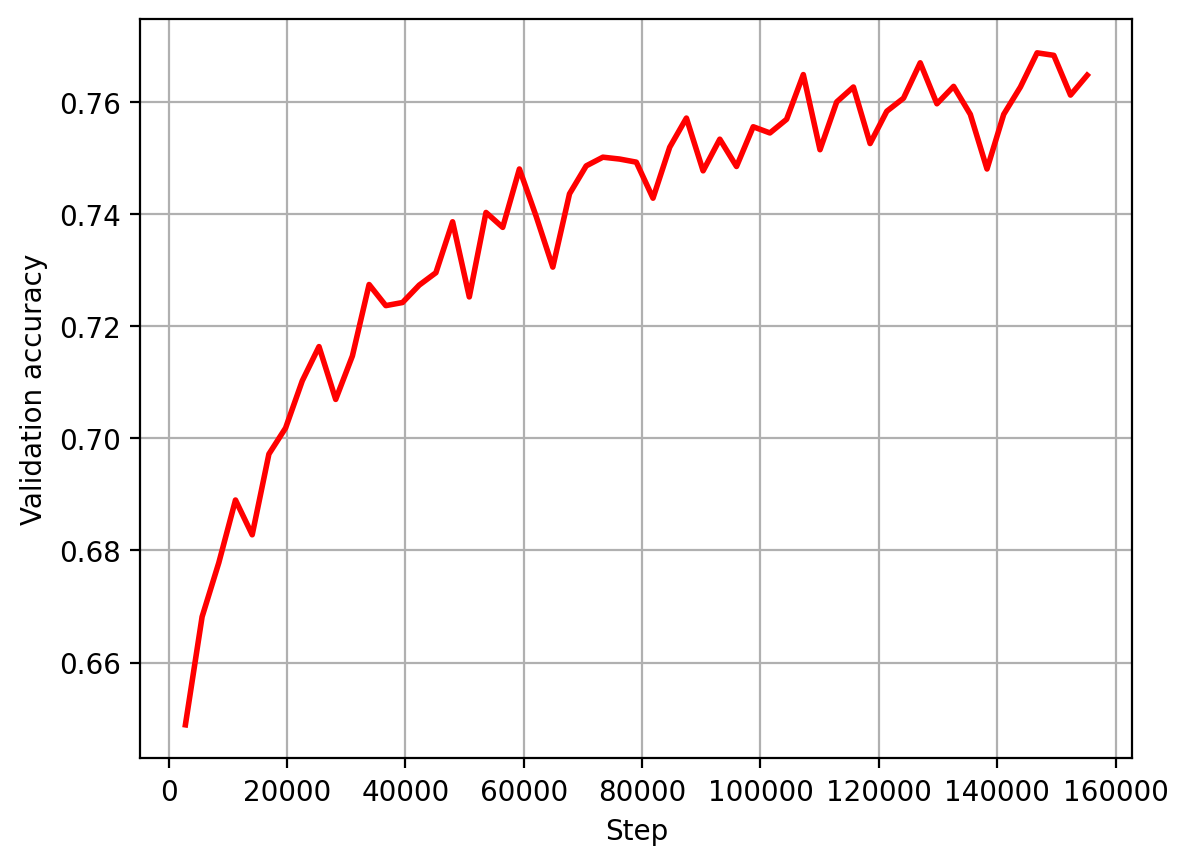}
    \caption{Validation plot}
    \label{fig:pick}
\end{figure}

\begin{figure}[htp]
    \centering
    \includegraphics[width=9cm]{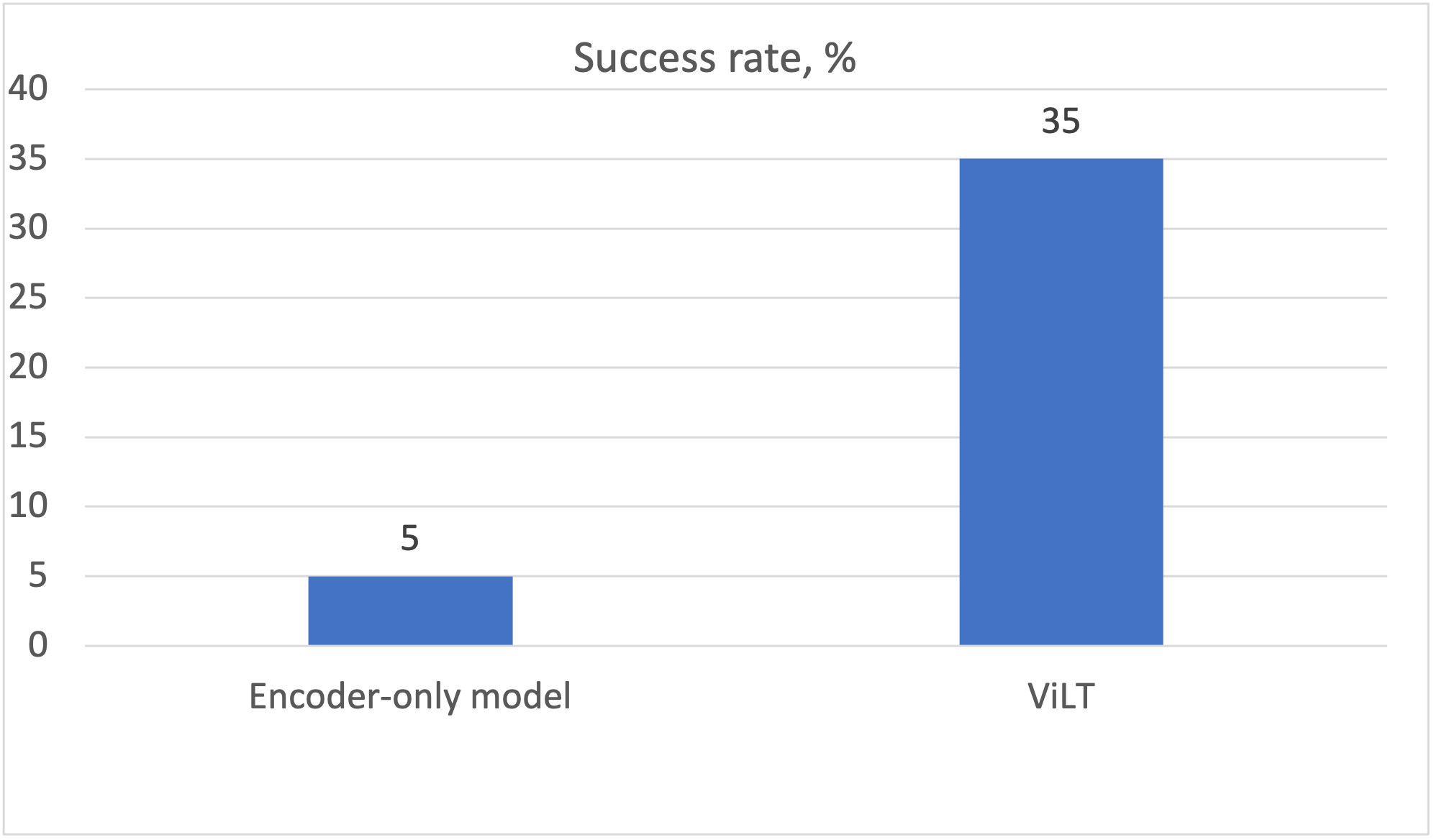}
    \caption{Success rate}
    \label{fig:plot_pick}
\end{figure}

For this experiment, 600 recordings of video from the robot's camera and the actions it performs were collected. The video recordings were spun at 25 frames per second and 160 x 120 px resolution. Each sample consists of 1000-1500 frames and an action number for each frame. Successful completion of a task is defined as receiving a command from the operator, finding the object to be moved, grabbing the object, finding the destination point, moving and stopping near or inside it. Results of validation during learning are presented in Fig. \ref{fig:pick}.
As can be seen from the graph, in this case the model also managed to find semantic dependencies, but despite the larger number of epochs and larger dataset the accuracy did not reach the desired value. 20 tests with different objects were conducted, the final result is shown on the graph (Fig. \ref{fig:plot_pick}).
In this task, the first model performed poorly because its generalization ability is insufficient for complex actions and long time series. The second model showed worse results than in the first experiment, this may be due to the small size of the dataset and its poor quality.

\subsection{Experiment 3, multitask}
\begin{figure}[htp]
    \centering
    \includegraphics[width=9cm]{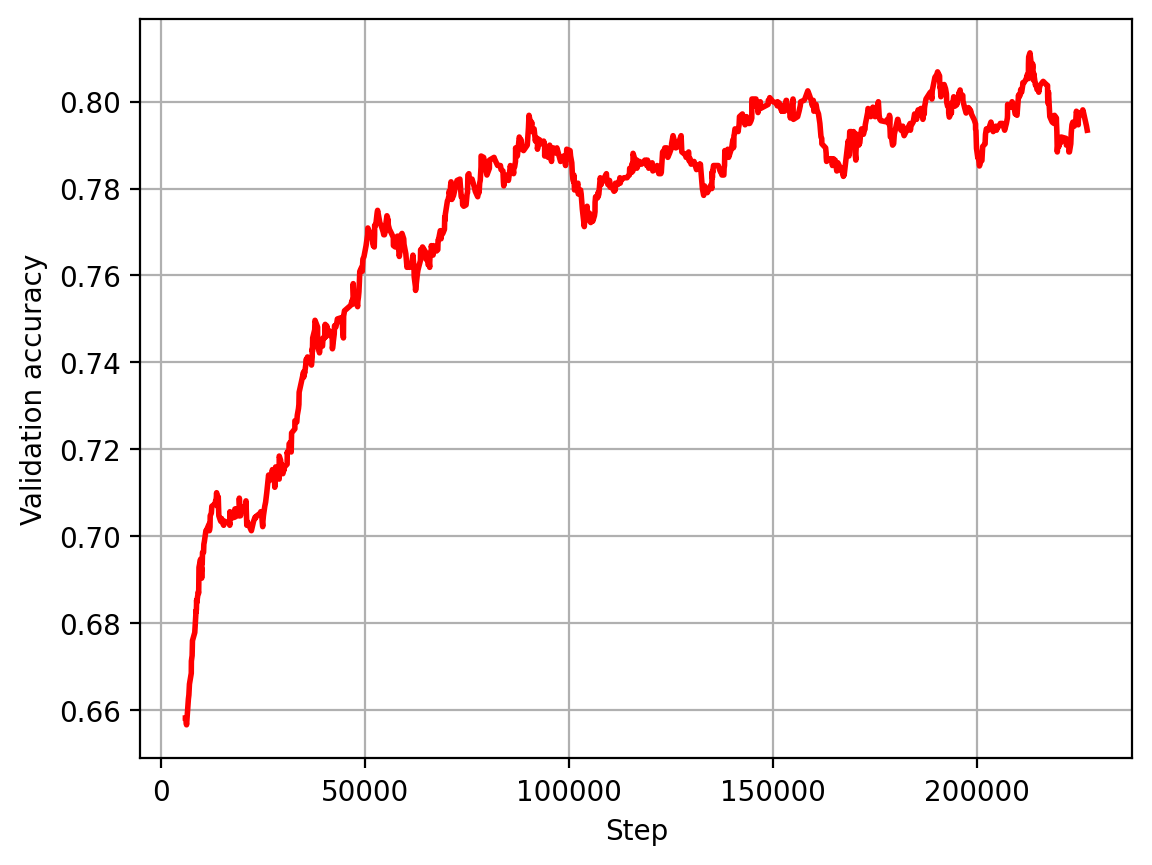}
    \caption{Validation plot}
    \label{fig:both}
\end{figure}

For this experiment was chosen the ViLT model due to its better performance on complex “Pick and place” task. This model was trained on mixed dataset, results validation during learning are presented in Fig. \ref{fig:both}.
As in the previous experiment, a limit is observed in the graph, this is due to the small size of the dataset and its lack of diversity. Nevertheless, the overall success rate from the 20 experiments was 45\%: 60\% for the “Go to point” task and 30\% for the “Pick and place” task. In this case, the result for the first task was higher than in the first experiment, where the model was trained only on the data from this task. From this it can be concluded that the model can generalize knowledge about more complex tasks and use it for smaller and simpler ones.

\section{Conclusion}

The thesis work proposes an approach to mission planning that is to abandon classical navigation and localization algorithms and use neural network models based on the Transformer architecture pretrained on web-scale data. The study proposed two approaches to the creation of such models: classification and generative. During the experiments it was found out that encoder-only approach is suitable only for simple tasks, such as point-to-point tracking, for more complex ones a generative model with more powerful generalization ability is needed. The work achieved a success rate of 53\% for “Go to point” task, 35\% for “Pick and place” task and 45\% for multitask. These results suggest that the approach has prospects for development, but a larger and better dataset is needed for further research. At this stage, we can conclude that the hypothesis that it is possible to abandon classical robotics algorithms has been confirmed.

\section*{Acknowledgements} 
Research reported in this publication was financially supported by the RSF grant No. 24-41-02039.


\begin{thebibliography}{00}
\bibitem{b1} Dosoftei, C.C. “Hardware in the Loop Topology for an Omnidirectional Mobile Robot Using Matlab in a Robot Operating System Environment”. Symmetry 2021, 13, 969.
\bibitem{b2} Pütz S. “Continuous shortest path vector field navigation on 3d triangular meshes for mobile robots”. In 2021 IEEE International Conference on Robotics and Automation.
\bibitem{b3} Salem I. E. “Flight-schedule using Dijkstra's algorithm with comparison of routes findings”. International Journal of Electrical and Computer Engineering, 12(2), p. 1675, 2022.
\bibitem{b4} Lu X “Design of adaptive sliding mode controller for four-Mecanum wheel mobile robot”. In Proceedings of the 2018 37th Chinese Control Conference (CCC), Wuhan, China, 25–27 July 2018; IEEE: Piscataway, NJ, USA, 2018; pp. 3983–3987.
\bibitem{b5} Ngwenya T. “Virtual Obstacles for Sensors Incapacitation in Robot Navigation: A Systematic Review of 2D Path Planning”. Sensors 2022, 22, 6943.
\bibitem{b6} Z. Hu et al., “Deploying and Evaluating LLMs to Program Service Mobile Robots”. In IEEE Robotics and Automation Letters, vol. 9, no. 3, pp. 2853-2860, March 2024.
\bibitem{b7} Wang Z. “Hybrid offline and online task planning for service robot using object-level semantic map and probabilistic inference”. Inf. Sci. 2022, 593, 78–98.
\bibitem{b8} Kannan “Smart-llm: Smart multi-agent robot task planning using large language models” arXiv preprint arXiv:2309.10062 (2023).
\bibitem{b9} Goyal et al. “Rvt: Robotic view transformer for 3d object manipulation”. Conference on Robot Learning. PMLR, 2023.
\bibitem{b10} Z. Hu et al., “Deploying and Evaluating LLMs to Program Service Mobile Robots”. In IEEE Robotics and Automation Letters, vol. 9, no. 3, pp. 2853-2860, March 2024.
\bibitem{b11} Brohan, Anthony, et al. “Rt-1: Robotics transformer for real-world control at scale” arXiv preprint arXiv:2212.06817 (2022).
\bibitem{b12} Michael Ahn et al. “Do as I can, not as I say: Grounding language in robotic affordances” arXiv preprint arXiv:2204.01691, 2022.
\bibitem{b13} Kuan Fang “Scene memory transformer for embodied agents in long-horizon tasks”. In Proceedings of the IEEE/CVF Conference on Computer Vision and Pattern Recognition, pp. 538–547, 2019.
\bibitem{b14} Mingxing Tan and Quoc Le. EfficientNet: Rethinking model scaling for convolutional neural net- works. In Kamalika Chaudhuri and Ruslan Salakhutdinov (eds.), Proceedings of the 36th In- ternational Conference on Machine Learning, volume 97 of Proceedings of Machine Learning Research, pp. 6105–6114. PMLR, 09–15 Jun 2019.
\bibitem{b15} Brohan et al. “Rt-2: Vision-language-action models transfer web knowledge to robotic control”. arXiv preprint arXiv:2307.15818 (2023).
\bibitem{b16} X. Chen. Pali-x: On scaling up a multilingual vision and language model, 2023a.
\bibitem{b17} Z. J. Cui, et al. From play to policy: Conditional behavior generation from uncurated robot data. arXiv preprint arXiv:2210.10047, 2022.
\bibitem{b18} Chen, Lin, et al. "Transformer-based imitative reinforcement learning for multi-robot path planning." IEEE Transactions on Industrial Informatics (2023).
\bibitem{b19} B.Graham “Levit: a vision transformer in convnet’s clothing for faster inference,” in Proceedings of the IEEE/CVF International Conference on Computer Vision, pp. 12 259–12 269, 2021.
\bibitem{b20} Z.Ma “Learning selective communication for multi-agent path finding," IEEE Robotics and Automation Letters, vol. 7,  no. 2, pp. 1455-1462, Apr. 2022. 
\bibitem{b21} Shridhar, at al.  "Perceiver-actor: A multi-task transformer for robotic manipulation." Conference on Robot Learning. PMLR, 2023.
\end{thebibliography}
\end{document}